\documentclass{bmvc2k}
\pdfoutput=1

\title{Occlusion-Aware Object Localization, Segmentation and Pose Estimation}

\addauthor{Samarth Brahmbhatt}{http://www.cc.gatech.edu/~sbrahmbh}{1}
\addauthor{Heni Ben Amor}{http://henibenamor.weebly.com}{1}
\addauthor{Henrik Christensen}{http://www.cc.gatech.edu/~hic}{1}

\addinstitution{
 School of Interactive Computing\\
 Georgia Institute of Technology\\
 Atlanta, GA USA
}

\runninghead{Brahmbhatt \bmvaEtAl}{Occlusion-Aware Object Detection}

\DeclareMathOperator*{\argmin}{\arg\!\min}

\usepackage{amsfonts}
\usepackage{amssymb}

\usepackage{algorithm}
\usepackage{algorithmic}

\usepackage{placeins}
\graphicspath{{images/}{images/supplementary/}}
\DeclareGraphicsExtensions{.jpg,.png,.pdf,.eps}

\usepackage{multicol}
\usepackage{nameref}
\usepackage{zref-xr}
\zxrsetup{toltxlabel}

\begin{document}

\maketitle

\begin{abstract}
We present a learning approach for localization and segmentation of objects in an image in a manner that is robust to partial occlusion. Our algorithm produces a bounding box around the full extent of the object and labels pixels in the interior that belong to the object. Like existing segmentation aware detection approaches, we learn an appearance model of the object and consider regions that do not fit this model as potential occlusions. However, in addition to the established use of pairwise potentials for encouraging local consistency, we use higher order potentials which capture information at the level of image segments. We also propose an efficient loss function that targets both localization and segmentation performance. Our algorithm achieves 13.52\% segmentation error and 0.81 area under the false-positive per image vs. recall curve on average over the challenging CMU Kitchen Occlusion Dataset. This is a 42.44\% decrease in segmentation error and a 16.13\% increase in localization performance compared to the state-of-the-art. Finally, we show that the visibility labelling produced by our algorithm can make full 3D pose estimation from a single image robust to occlusion.
\end{abstract}

\section{Introduction and Related Work} \label{sec:1}
In this paper we address the problem of localizing and segmenting partially occluded objects. We do this by generating a bounding box around the full extent of the objects, while also segmenting the visible parts inside the box. This is different from semantic segmentation, which typically does not provide information about the spatial position of labelled pixels inside the object. While a lot of progress has been made in object detection~\cite{dpm_main, line2d, menglong}, occlusion by other objects still remains a challenge. A common theme is to model occlusion geometrically or appearance-wise, thereby allowing it to contribute to the detection process.~\citet{wang_hog} use a holistic Histogram of Oriented Gradients (HOG) template~\cite{hog_main} to scan through the image and use specially trained part templates for instances where some cells of the holistic template respond poorly.~\citet{rbg_grammar} force the Deformable Part Model detector to place a trained `occluder' part in regions where the original parts respond weakly. The object masks produced by both of these algorithms are only accurate up to the parts and hence not usable for many applications e.g. edge-based 3D pose estimation.~\citet{sav_3Dasp} approximate object structure in 3D using planar parts. A Conditional Random Field (CRF) is then used to reason about visibility of the parts when the 3D planes are projected to the image. However, such methods work well only for large objects that can be approximated with planar parts.\par

Our approach is entitled \emph{Segmentation and Detection using Higher-Order Potentials} (SD-HOP). It is based on discriminatively learned HOG templates for objects and occlusion. Whereas the object templates model the objects of interest, the occlusion templates provide discriminative support and do not model a specific occluder. Segmentation is done by considering not only the response of patches to these templates, but also the segmentation of neighbouring patches through a CRF with higher-order connections that encompass image regions.\par

We will compare our approach to two existing approaches that have been designed to handle partial occlusion.~\citet{cmu_occlusion} approximate all occluders by boxes and generate occlusion hypotheses by finding locations of mismatch between image gradient and object model gradient. These hypotheses are then validated by the visibility of other points of the object and by an occlusion prior which assumes all objects rest on the same planar surface. Our algorithm does not need such assumptions which reduce the segmentation accuracy.~\citet{segaware} learn discriminative appearance models of the object and occlusion seen during training. Segmentation is achieved by defining a CRF to assign binary labels to patches based on their response to these two filters. We build on their work but add several important modifications that lead to better localization and segmentation performance. Firstly, we replace the edge-based pairwise terms with 4-connected pairwise terms that are better able to propagate visibility relations. Secondly, we introduce the use of higher-order potentials defined over groups of patches, allowing us to reason at the level of image segments which contain much more information than pairs of patches. We also introduce a new loss function for structured learning that targets both localization and segmentation performance but is still decomposable over the energy terms. Lastly, we introduce a simple procedure to convert the granular patch-level object mask produced by the algorithm to a fine pixel-level mask that can be used to make 3D pose estimation of detected objects robust to partial occlusion. Our algorithm outperforms these approaches (\citet{cmu_occlusion},~\citet{segaware}) at both object localization and segmentation on the CMU Kitchen Occlusion dataset as shown in Section~\ref{sec:3}.\par

The rest of the paper is organized as follows. Section~\ref{sec:2} describes our proposed approach. We present evaluations on standard datasets and our own laboratory dataset in Section~\ref{sec:3} and summarize in Section~\ref{sec:4}.

\section{Method} \label{sec:2}
\begin{figure}
\centering
\includegraphics[width=\textwidth]{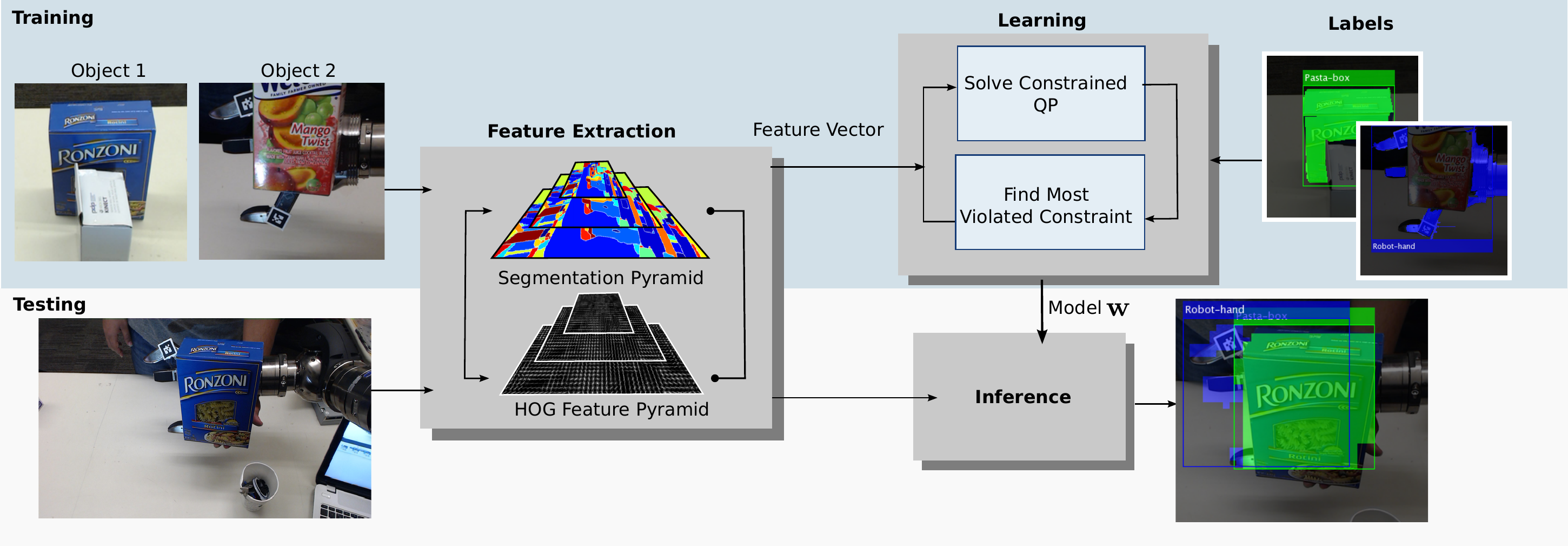}
\caption{Overview of our approach. \textit{Top}: During training, images are segmented and features are extracted from pyramids of segmentations and HOG features. An SVM model is learned by max-margin learning. \textit{Bottom}: After training, the model can be used to infer a bounding box and visible segments of the object.}
\label{fig:1}
\end{figure}
The training phase for SD-HOP requires a set of images with different occlusions of the object(s) of interest. Each training sample is (1) over-segmented and (2) annotated with a bounding box around the full extent of the object and a binary segmentation of the area inside the box into object vs. non-object pixels. Given these training images and labels, we train a structured Support Vector Machine (SVM) that  produces the HOG templates and CRF weights. Figure~\ref{fig:1} shows an overview of our approach.\par

Object segmentation is done by assigning binary labels to all HOG cells within the bounding box, 1 for visible and 0 for occluded. Instead of making independent decisions for every cell, we allow  neighbouring cells to influence each other. Neighbour influence can take two forms: (1) pairwise terms (\citet{grab_cut}) that impose a cost for 4-connected neighbours to have different labels and (2) higher-order potentials (\citet{kohli_hop}) that impose a cost for cells to have a different label than the dominant label in their segment of the image. These segments are produced separately by an unsupervised segmentation algorithm.

\subsection{Notation}
\noindent
The label for an object in an image \textbf{x} is represented as $\textbf{y} = (\textbf{p}, \textbf{v}, a)$, where $\textbf{p}$ is the bounding box, \textbf{v} is a vector of binary variables indicating the visibility of HOG cells within $\textbf{p}$ and $a \in [1, A]$ indexes the discrete viewpoint. $\mathbf{p} = (p_x, p_y, p_{\sigma})$ indicates the position of the top left corner and the level in a scale-space pyramid. The width and height of the box are fixed per viewpoint as $w_a$ and $h_a$ HOG cells respectively. Hence \textbf{v} has $w_a \cdot h_a$ elements. All training images are also over-segmented to collect statistics for higher-order potentials. Any unsupervised algorithm can be used for this, e.g.~\citet{fz_seg} and~\citet{gpb_ucm}.

\subsection{Feature Extraction}\label{subsec:features}
\noindent
Given an image $\mathbf{x}$ and a labelling $\mathbf{y}$, a sparse joint feature vector $\Psi(\mathbf{x},\mathbf{y})$ is formed by stacking $A$ vectors. Each of these vectors has features for a different discretized viewpoint. All vectors except for the one corresponding to viewpoint $a$ are zeroed out. Below, we describe the components of this vector.
\begin{enumerate}
\itemsep0em
\item 31-dimensional HOG features are extracted for all cells of 8x8 pixels in $\mathbf{p}$ as described in \citet{dpm_main}. The feature vector is is constructed by stacking two groups which are formed by zeroing out different parts, similarly to~\citet{vedaldi_structured}. The visible group $\phi_v(\mathbf{x},\textbf{p})$ has the HOG features zeroed out for cells labelled 0 and the occlusion group $\phi_{nv}(\mathbf{x},\textbf{p})$ has them zeroed out for cells labelled 1. 
\item Complemented visibility labels, to learn a prior for a cell to be labelled 0: $\left[\textbf{1}_{wh} - \textbf{v}\right]$.
\item Count $c(\mathbf{p})$ of cells in bounding box \textbf{p} lying outside the image boundaries, to learn a cost for truncation by the image boundary, similarly to~\citet{vedaldi_structured}.
\item Number of 4-connected neighbouring cells in the bounding box that have different labels, to learn a pairwise cost.
\item Each segment in the bounding box obtained from unsupervised segmentation defines a clique of cells. To learn higher-order potentials, we need a vector $\theta_{HOP}$ that captures the distribution of 0/1 label agreement within cliques. A vector $\theta_c \in \mathbb{R}^{K+1}$ is constructed for each clique $c$ as $(\theta_c)_k = 1$ if $\sum_{i \in c}v_i = k$. The sum of all $\theta_c$ within $\mathbf{p}$ gives $\theta_{HOP}$. In practice, since cliques do not have the same size we employ the normalization strategy described in~\citet{low_lin} and transform statistics of all cliques to a standard clique size $K$ ($K=4$ in our experiments).
\item The constant $1$, used to learn a bias term for different viewpoints.
\end{enumerate}

\subsection{Learning}
\noindent
Suppose $\mathbf{w}$ is a vector of weights for elements of the joint feature vector. We define $\mathbf{w}^T\Psi(\mathbf{x},\mathbf{y})$ as the `energy' of the labelling $\mathbf{y}$. The aim of learning is to find $\mathbf{w}$ such that the energy of the correct label is minimum. Hence we define the label predicted by the algorithm as
\begin{equation}\label{eq:3.1}
f(\mathbf{x}; \mathbf{w}) = \mathbf{y^*} = \argmin_{\mathbf{y}} \mathbf{w}^T\Psi(\mathbf{x},\mathbf{y})
\end{equation}
We use a labelled dataset $\left(\mathbf{x}_i, \mathbf{y}_i\right)_{i=1}^N$ and learn $\mathbf{w}$ by solving the following constrained Quadratic Program (QP)
\begin{equation}\label{eq:3.2}
\min_{\mathbf{w},\xi}\frac{1}{2}\|\textbf{w}\|_2+C\sum_{i=1}^{N}\xi_i
\end{equation}
\begin{align*}
\text{s.t.}\quad &\mathbf{w}^T (\Psi(\mathbf{x}_i, \hat{\mathbf{y}}_i) - \Psi(\mathbf{x}_i, \mathbf{y}_i)) + \xi_i \geq \Delta(\mathbf{y}_i, \hat{\mathbf{y}}_i)~\forall i, \hat{\mathbf{y}}\in Y_i\\
&\xi_i \geq 0~\forall i\\
&\mathbf{D}^2\mathbf{w} \geq \mathbf{0}
\end{align*}
Intuitively this formulation requires that the score $\mathbf{w}^T \Psi(\textbf{x}_i, \textbf{y}_i)$ of any ground truth labelled image $\textbf{x}_i$ must be smaller than the score $\mathbf{w}^T \Psi(\textbf{x}_i, \hat{\textbf{y}}_i)$ of any other labelling $\hat{\textbf{y}}_i$ by the distance between the two labellings $\Delta(\textbf{y}_i, \hat{\textbf{y}}_i)$ minus the slack variable $\xi_i$, where $\|\mathbf{w}\|_2$ and $\xi_i$ are minimized. The regularization constant $C$ adjusts the importance of minimizing the slack variables. The above formulation has exponential constraints for each training image. For tractability, training is performed by using the cutting plane training algorithm of~\citet{joa_svm} which maintains a working set $Y_i$ of most violated constraints (MVCs) for each image.~\citet{low_lin} adapts this algorithm for training higher-order potentials. It uses $\mathbf{D}^2$ as a second order curvature constraint on the $K+1$ weights for the higher-order potentials, which forces them to make a concave lower envelope. This encourages most cells in the image segments to agree in visibility labelling. $\mathbf{D}^2$ is an appropriately 0-padded (to the left and right) version of
\begin{equation*}
\begin{bmatrix}
-1 &  2 & 1 &  0 &\ldots\\
\vdots & \vdots & \ddots & \vdots\\
 0 & \ldots & -1 & 2 & -1 
\end{bmatrix}.
\end{equation*}
The distance between two labels $\mathbf{y}$ and $\mathbf{\hat{y}}$ is called the loss function. It depends on the amount of overlap between the two bounding boxes and the Hamming distance between the visibility labellings
\begin{equation}\label{eq:3.3}
\Delta(\textbf{y}, \hat{\textbf{y}}) = \left(1-\frac{\text{area}(\mathbf{p} \bigcap \hat{\mathbf{p}})}{\text{area}(\mathbf{p} \bigcup \hat{\mathbf{p}})}\right)+\frac{\text{area}(\mathbf{p} \bigcap \hat{\mathbf{p}})}{\text{area}(\mathbf{p} \bigcup \hat{\mathbf{p}})}\cdot H(\mathbf{v}, \hat{\mathbf{v}})
\end{equation}
The mean Hamming distance $H(\mathbf{v}, \hat{\mathbf{v}})$ between two labellings $\mathbf{v}$ and $\hat{\mathbf{v}}$ (potentially having different sizes as they might belong to different viewpoints) is calculated after projecting them to the lowest level of the pyramid. By construction of the loss function, the difference in segmentation starts contributing to the loss only after the two bounding boxes start overlapping each other. It also has the nice property of decomposing over the energy terms, as described in Section~\ref{subsubsec:mvc}.

\subsection{Inference}\label{subsec:inference}
\noindent
To perform the inference as described in Eq.~\ref{eq:3.1} we have to search through $Y=A \times P \times V$ where $A$ is the set of viewpoints, $P$ is the set of all pyramid locations and $V$ is the exponential set of all combinations of visibility variables. We enumerate over $A$ and $P$ and use an $s-t$ mincut to search over $V$ at every location.\par
\begin{figure}[t]
\centering
\subfigure[]{\label{fig:2a}\includegraphics[width=0.45\textwidth]{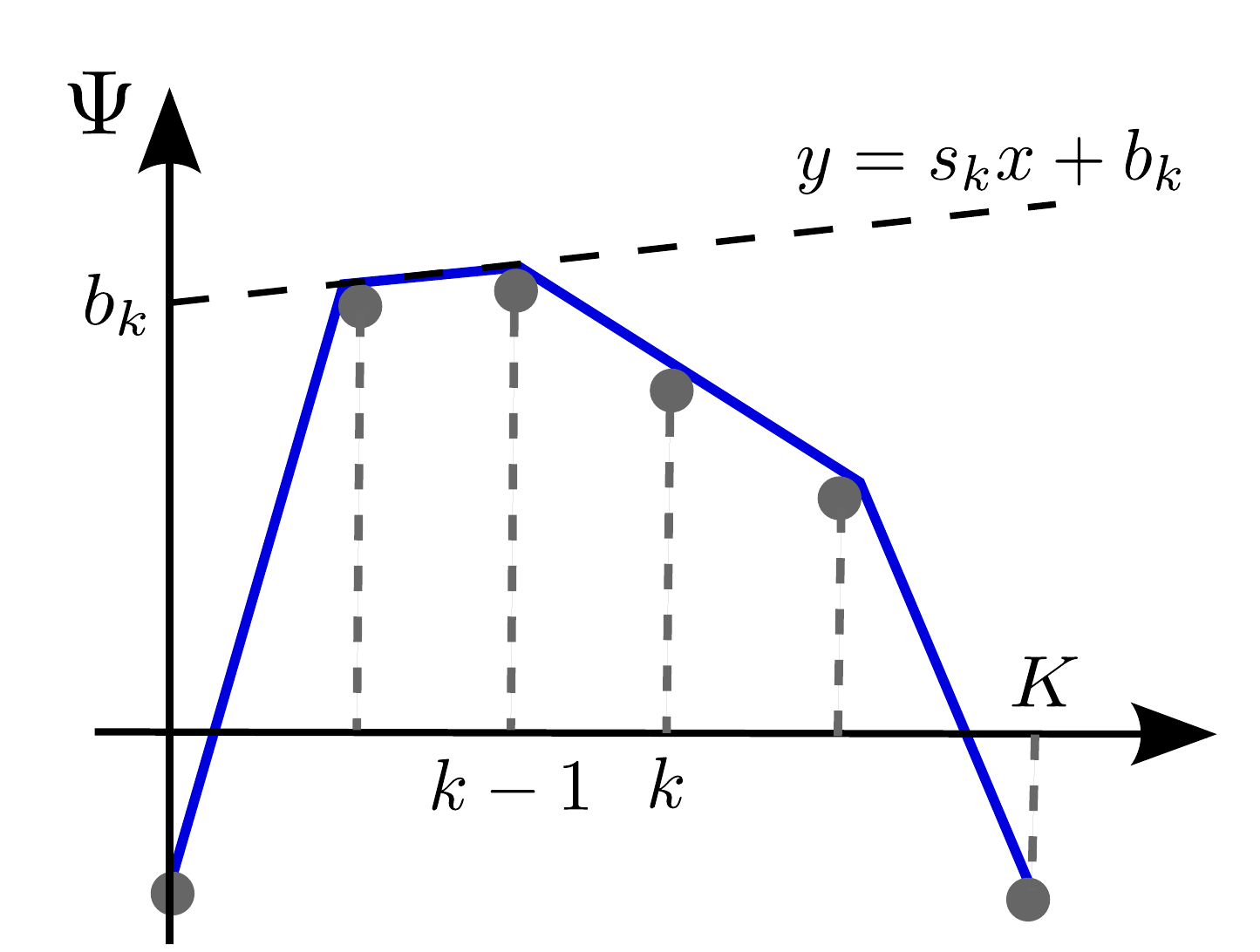}}
\subfigure[]{\label{fig:2b}\includegraphics[width=0.45\textwidth]{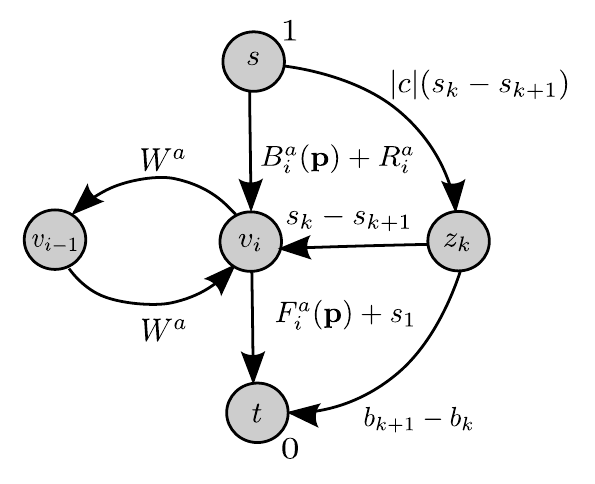}}
\caption{(a): Concave higher-order potentials encouraging cells in a clique to have the same binary label, (b): Construction of graph to compute the energy minimizing binary labelling of cells by $s-t$ mincut.}
\label{fig:2}
\end{figure}

By construction, the feature vector \textbf{w} can be decomposed into weight vectors for the different viewpoints i.e. $\mathbf{w} = [\mathbf{w}^1, \mathbf{w}^2, \ldots, \mathbf{w}^A]$. In the following description, we will consider one viewpoint and omit the superscript for brevity of notation. $\mathbf{w}$ can also be decomposed as $[\mathbf{w}_v, \mathbf{w}_{nv}, \mathbf{w}_{pr}, w_{trunc}, W, \mathbf{w}_{HOP}, w_{bias}]$ into the six components described in Section~\ref{subsec:features}. We define the following terms that are used to construct the graph shown in Figure~\ref{fig:2b}. $\phi_i(\mathbf{x},\mathbf{p})$ are the vectorized HOG features extracted at cell $i$ in bounding box $\mathbf{p}$. Unary terms $F_i(\mathbf{p})={\mathbf{w}_{v,i}}^T\phi_i(\mathbf{x},\mathbf{p})$ and $B_i(\mathbf{p})={\mathbf{w}_{nv,i}}^T\phi_i(\mathbf{x},\mathbf{p})$ are the responses at cell $i$ for object and occlusion filters respectively. $R_i = \mathbf{w}_{pr, i}$ is the prior for cell $i$ to be labelled 0. Constant term $C(\mathbf{y}) = w_{trunc} \cdot c(\mathbf{p}) + w_{bias}$ is the sum of image boundary truncation cost and bias. $\mathcal{E}$ is the set of 4-connected neighbouring cells in $\mathbf{p}$ and $W$ is the pairwise weight. $\mathcal{C}(\mathbf{p})$ is the set of all cliques in $\mathbf{p}$ and $\psi_c(\mathbf{v}_c)$ is the higher-order potential for clique $c$ having nodes with visibility labels $\mathbf{v}_c$. Combining these terms, the energy for a particular labelling is formulated as
\begin{align}\label{eq:3.4}
\begin{aligned}
E(\mathbf{x}, \mathbf{y}) = \textbf{w}^{T} \Psi(\textbf{x}, \textbf{y})
&= \sum_{i=1}^{wh}F_i(\mathbf{p})v_i+B_i(\mathbf{p})(1-v_i)+R_i(1-v_i)\\
&+\sum_{(i,j) \in \mathcal{E}}W|v_i-v_j|+\sum_{c \in \mathcal{C}(\mathbf{p})}\psi_c(\mathbf{v}_c) + C(\mathbf{y})
\end{aligned}
\end{align}
$\psi_c(\mathbf{v}_c)$, the higher-order potential for clique $c$ is defined as $\min_{k = 1 \ldots K}\left(s_k \sum_{i \in c}{v_i} + b_k\right)$, following~\citet{low_lin}. Intuitively, it is the lower envelope of a set of lines whose slope is defined as $s_k=\frac{M}{K}\left((w_{HOP})_k-(w_{HOP})_{k-1}\right)$ and intercept as $b_k=(w_{HOP})_k-s_kk$ (recall that $\mathbf{w}_{HOP}$ is a $K+1$ dimensional weight vector). $M$ is the size of the clique. The normalization in $s_k$ makes the potential invariant to the size of the clique (refer to~\citet{low_lin} for details). Figure~\ref{fig:2a} shows a sample higher-order potential curve for a clique of $K$ cells.\par

Given an image, a location, and a viewpoint we use $s-t$ mincut on the graph construction shown in Figure~\ref{fig:2b} to find the labelling $\mathbf{v}$ that minimizes the energy in Eq.~\ref{eq:3.4}. Each variable $v_i$, $i \in \lbrace 1,2,\ldots ,wh\rbrace$ defines a node and each clique has $K-1$ auxiliary nodes in the graph, $z_1 \ldots z_{K-1}$. For a detailed derivation of this graph structure please see~\citet{kol_gc} and~\citet{low_lin}.
After the maxflow algorithm finishes, the nodes $v_i$ still connected to $s$ are labelled $1$ and others are labelled $0$.

\subsubsection{Loss-augmented Inference}\label{subsubsec:mvc}
Loss-augmented inference is an important part of the cutting plane training algorithm (`separation oracle' in~\citet{joa_svm}) and is used to find the most violated constraints. It is defined as $\mathbf{y_{MVC}} = \argmin_{\mathbf{\hat{y}}} \mathbf{w}^T\Psi(\mathbf{x},\mathbf{\hat{y}}) - \Delta(\mathbf{y}, \mathbf{\hat{y}})$, where $\mathbf{y}$ is the ground-truth labelling. Our formulation of the loss function makes it solvable with the same complexity as normal inference (Eq.~\ref{eq:3.1}) by decomposing the loss over the terms in Eq~\ref{eq:3.4}. The first term of Eq.~\ref{eq:3.3} is added to $C(\mathbf{y})$, while the second term is distributed across $F_i(\mathbf{p})$ and $B_i(\mathbf{p})$ in Eq.~\ref{eq:3.4}.

\subsection{Detection of Multiple Objects} \label{subsec:multiple}
\noindent
Multiple objects of interest might overlap. Running the individual object detectors separately leaves regions of ambiguity in overlapping areas if multiple detectors mark the same location as visible. We find that running one iteration of $\alpha$-expansion (see~\citet{boy_ae}) in overlapping areas resolves ambiguities coherently. The detectors are run sequentially. We maintain a label map $\mathcal{V}$ that stores for each cell the label of the object that last marked it visible, and a collected response map $C$ that stores for each cell the object filter response ($F_i(\mathbf{p})$) from the object that last marked it visible. While running the location search for object $o$, we transfer object filter responses from $C$ to the occlusion filter response map ($B(\mathbf{p})$) for the current object as described in Algorithm~\ref{alg:multiple}.
\begin{algorithm}[t]
\caption{Response-transfer between object detectors in overlapping regions}\label{alg:multiple}
\begin{algorithmic}
\FORALL[$L$ is the number of objects, $\circ$ denotes the Hadamard product]{$o \in [1, L]$}
\FORALL{$\mathbf{p} \in P$}
\STATE{$B(\mathbf{p})=C(\mathbf{p}) \circ \mathbf{1}[\mathcal{V}(\mathbf{p}) \neq 0]$} \COMMENT{Transfer equation for all cells in $\mathbf{p}$}
\ENDFOR
\STATE{$\mathbf{y^*} = \argmin_{\mathbf{y}} \mathbf{w}^T\Psi(\mathbf{x},\mathbf{y})$}
\STATE{$C(\mathbf{y^*}(\mathbf{p})) = F(\mathbf{y^*}(\mathbf{p})) \circ \mathbf{y^*}(\mathbf{v})$} \COMMENT{Update equations for all cells in $\mathbf{p}$}
\STATE{$\mathcal{V}(\mathbf{y^*}(\mathbf{p})) = o \cdot \mathbf{y^*}(\mathbf{v})$}
\ENDFOR
\end{algorithmic}
\end{algorithm}
This is effectively one iteration of $\alpha$-expansion (see supplementary material for details). It causes decisions in overlapping regions to be made between responses of well-defined object filters rather than between responses of an object filter and a generic occlusion filter.\par

Such response-transfer requires the object models to be compatible with each other. We achieve this by training the object models together as if they were different viewpoint components of the same object. The bias term in the feature vector makes the filter responses of different components comparable.
 
\subsection{3D Pose Estimation}
\noindent
The basic principle of many model based 3D pose estimation algorithms is to fit a given 3D model of the object to its corresponding edges in the image e.g. in~\citet{choi_pose}, the 3D CAD model is projected into the image and correspondences between the projected model edges and image edges are set up. The pose is estimated by solving an Iterative Re-weighted Least Squares (IRLS) problem. However, partial occlusion causes these approaches to fail by introducing new edges. We make the algorithm robust to partial occlusion by first identifying visible pixels of the object using SD-HOP and discarding correspondences outside the visibility mask.  We call our extension of the algorithm Occlusion Reasoning-IRLS (OR-IRLS).

\section{Evaluation} \label{sec:3}
We implemented SD-HOP in Matlab, with MVC search and inference implemented in CUDA since they are massively parallel problems. Inference on a 640x480 image with 11 scales takes 3s for a single object with a single viewpoint on our 3.4 GHz CPU and NVIDIA GT-730 GPU.
 
\subsection{Localization and Segmentation}
\begin{figure}[!t]
\centering
\subfigure{\includegraphics[trim=0 5 0 7,clip,width=0.24\textwidth]{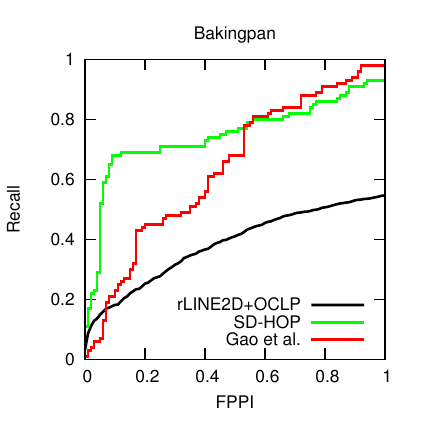}}
\subfigure{\includegraphics[trim=0 5 0 7,clip,width=0.24\textwidth]{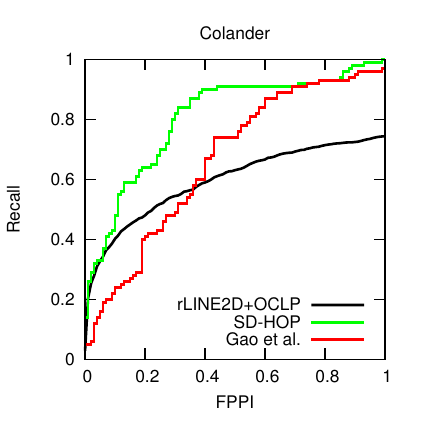}}
\subfigure{\includegraphics[trim=0 5 0 7,clip,width=0.24\textwidth]{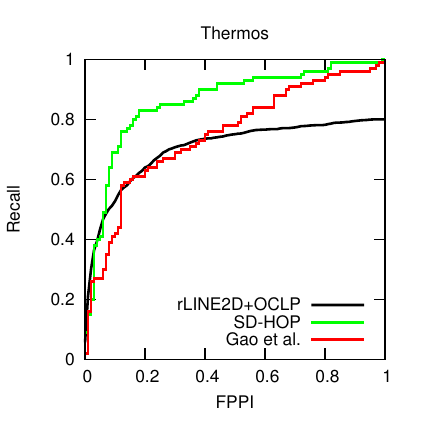}}
\subfigure{\includegraphics[trim=0 5 0 7,clip,width=0.24\textwidth]{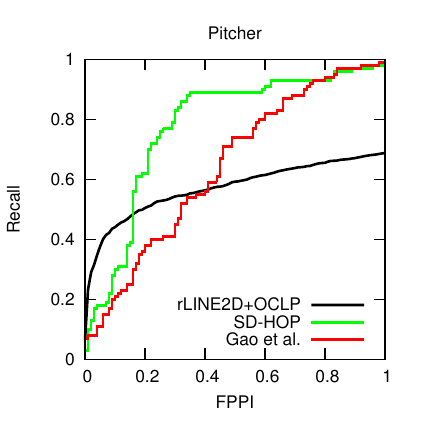}}
\subfigure{\includegraphics[trim=0 5 0 7,clip,width=0.24\textwidth]{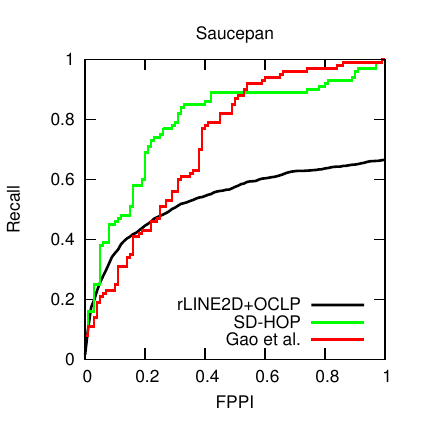}}
\subfigure{\includegraphics[trim=0 5 0 7,clip,width=0.24\textwidth]{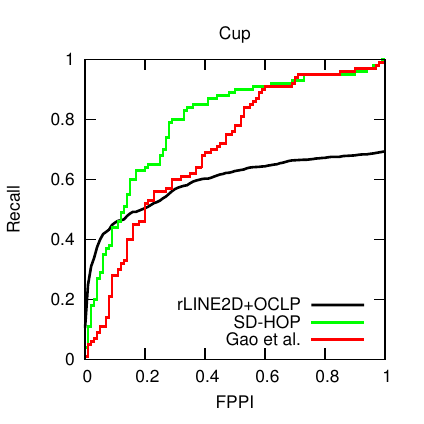}}
\subfigure{\includegraphics[trim=0 5 0 7,clip,width=0.24\textwidth]{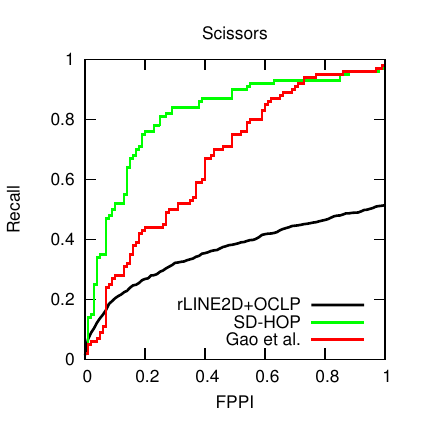}}
\subfigure{\includegraphics[trim=0 5 0 7,clip,width=0.24\textwidth]{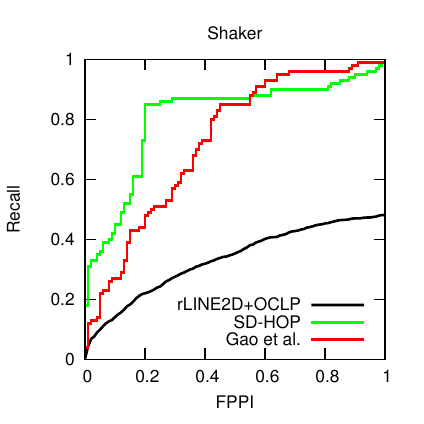}}
\caption{Object localization results on the CMU Kitchen Occlusion dataset}
\label{fig:3}
\end{figure}

We evaluated our approach on the CMU Kitchen Occlusion Dataset from~\citet{cmu_occlusion}. This dataset was chosen because (1) it provides extensive labelled training data in the form of images with bounding boxes and object masks, and (2) the dataset is challenging and offers the opportunity to compare against an algorithm designed specifically to handle occlusion. For the localization task we generated false positives per image (FPPI) vs. recall curves, while for the segmentation task we measured the mean segmentation error against ground truth as defined by the Pascal VOC segmentation challenge in ~\citet{pas_voc}. $C = 25$ (see eq.~\ref{eq:3.2}) was chosen by 5-fold cross-validation. While both results are presented for the single pose part of the dataset, multiple poses are easily handled in our algorithm as different components of the feature vector. Figure~\ref{fig:3} shows FPPI vs. recall curves compared with those reported by the rLINE2d+OCLP algorithm of~\citet{cmu_occlusion} and those generated from our implementation of ~\citet{segaware}. Table~\ref{table:1} presents segmentation errors compared with~\citet{segaware}.~\citet{cmu_occlusion} do not report a segmentation of the object.

\begin{figure}[!t]
\centering
\subfigure{\includegraphics[width=0.3\textwidth]{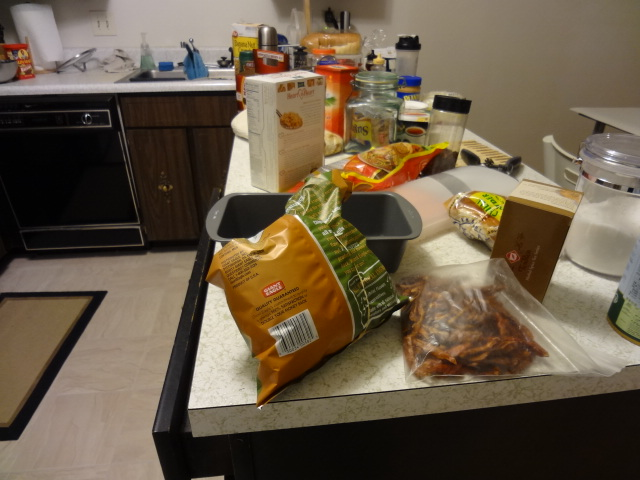}}
\subfigure{\includegraphics[width=0.3\textwidth]{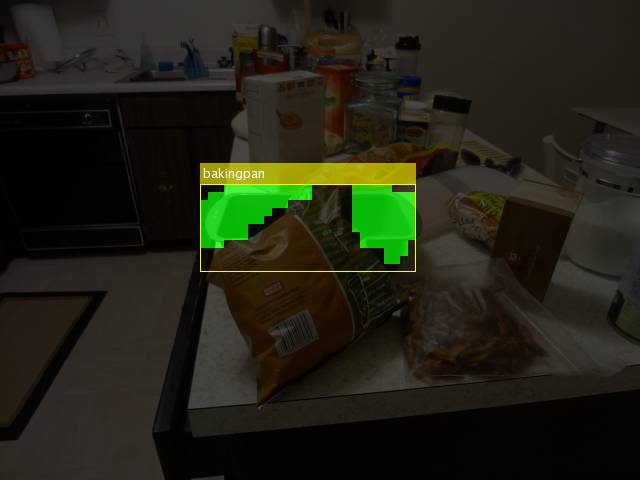}}
\subfigure{\includegraphics[width=0.3\textwidth]{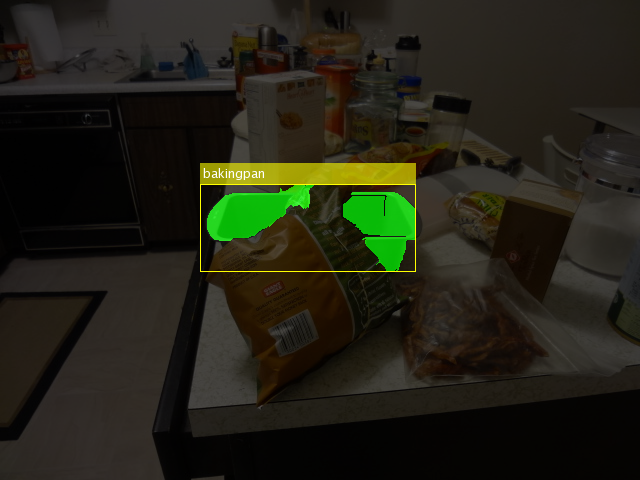}}\\[-2ex]
\subfigure{\includegraphics[width=0.3\textwidth]{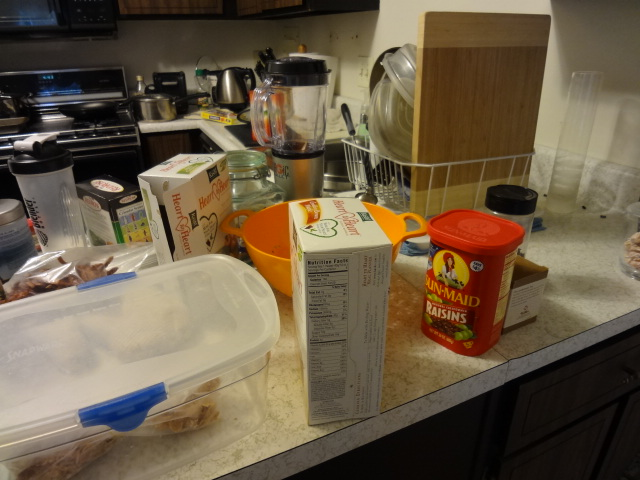}}
\subfigure{\includegraphics[width=0.3\textwidth]{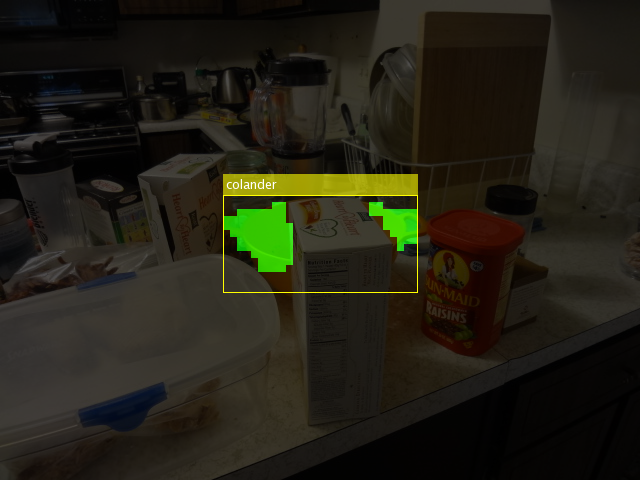}}
\subfigure{\includegraphics[width=0.3\textwidth]{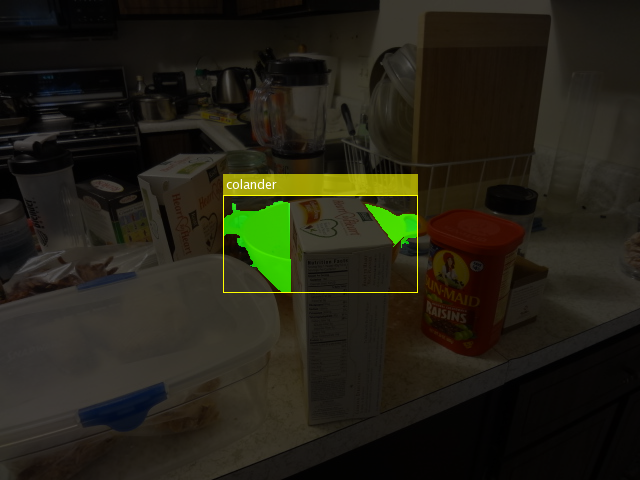}}\\[-2ex]
\subfigure{\includegraphics[width=0.3\textwidth]{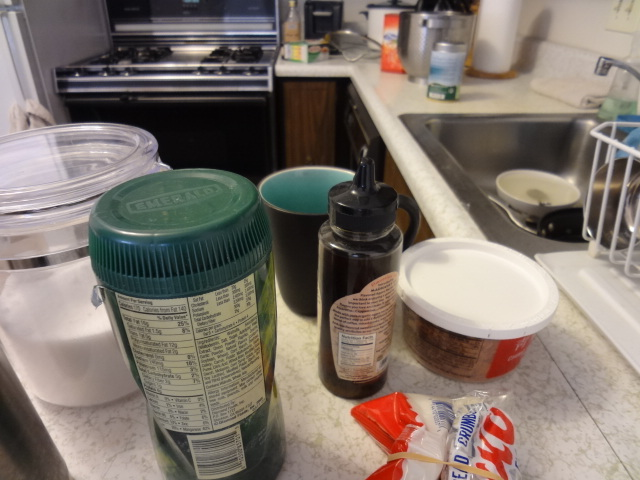}}
\subfigure{\includegraphics[width=0.3\textwidth]{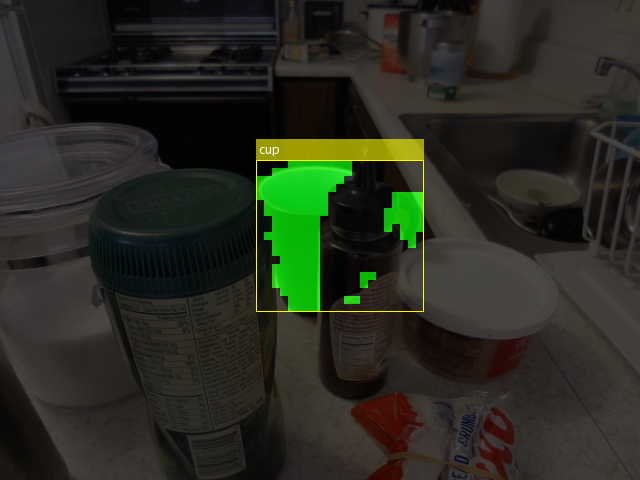}}
\subfigure{\includegraphics[width=0.3\textwidth]{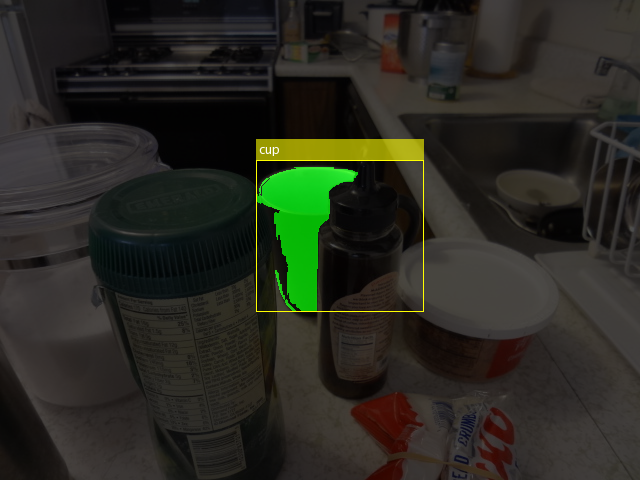}}\\[-2ex]
\subfigure{\includegraphics[width=0.3\textwidth]{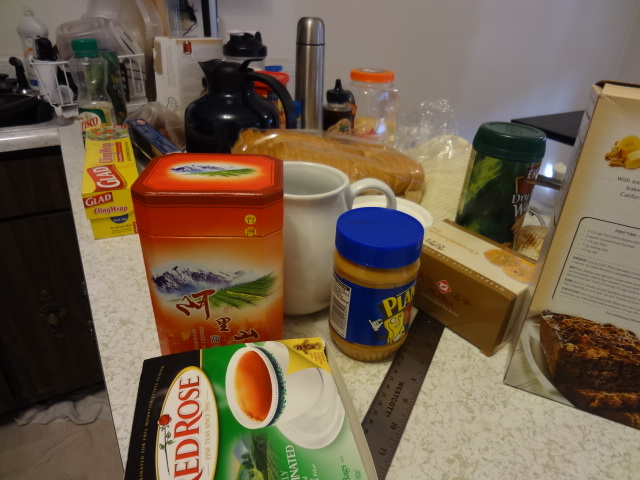}}
\subfigure{\includegraphics[width=0.3\textwidth]{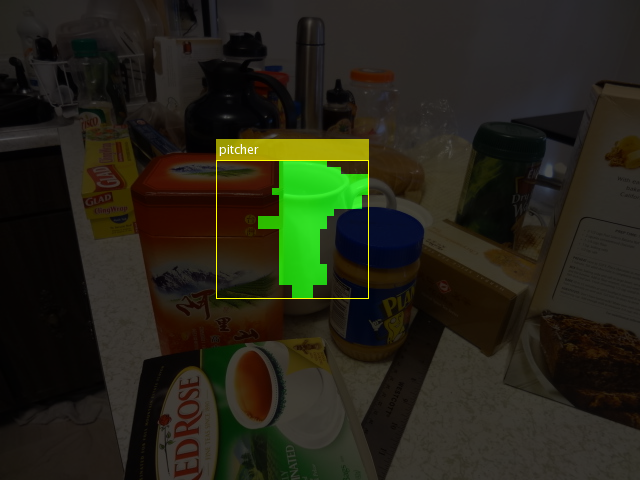}}
\subfigure{\includegraphics[width=0.3\textwidth]{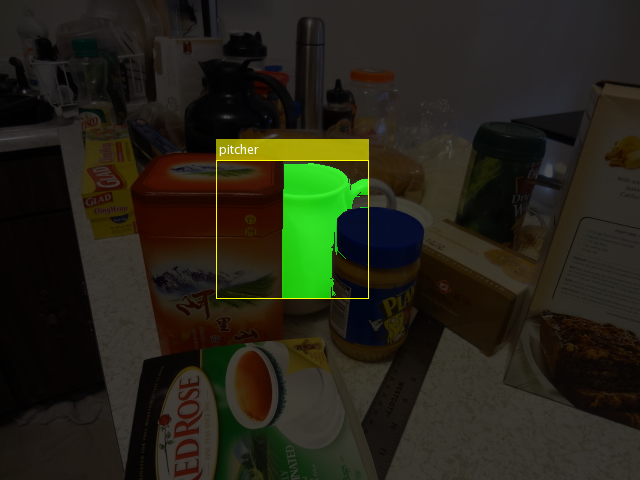}}\\[-2ex]
\subfigure{\includegraphics[width=0.3\textwidth]{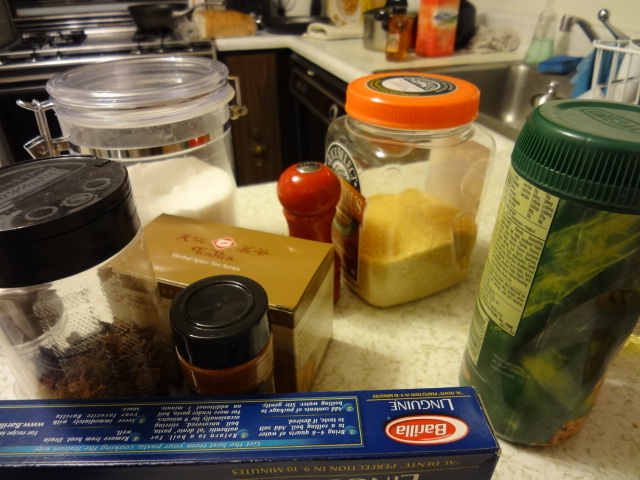}}
\subfigure{\includegraphics[width=0.3\textwidth]{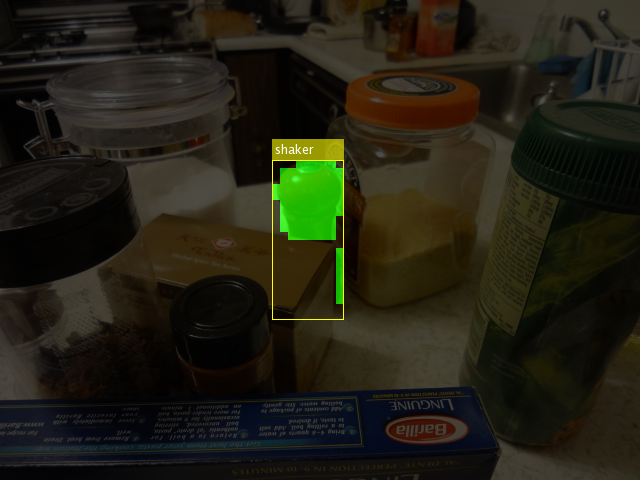}}
\subfigure{\includegraphics[width=0.3\textwidth]{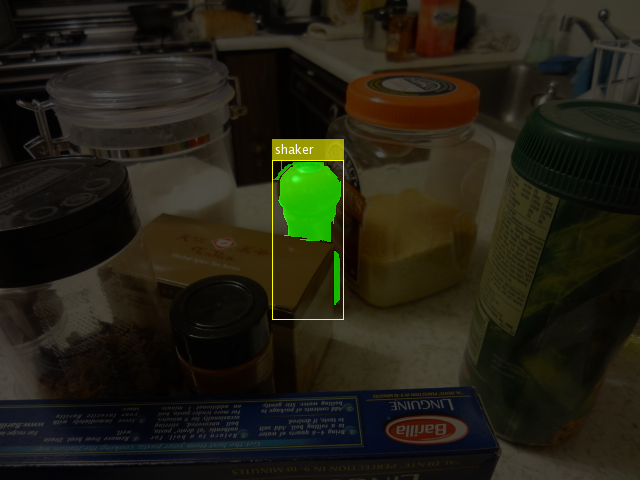}}
\caption{Object localization and segmentation results on the CMU Kitchen Occlusion dataset. Left: Image, Center: Raw mask from SD-HOP, Right: Refined mask from SD-HOP}
\label{fig:5}
\end{figure}

\begin{table}[h]
\renewcommand{\arraystretch}{1.3}
\parbox{0.5\linewidth} {
\caption{Mean object segmentation error}
\label{table:1}
\centering
\begin{tabular}{*3c}
\hline
\textbf{Object} & \textbf{\citet{segaware}} & \textbf{SD-HOP}\\ \hline
Bakingpan & 0.2904 & \textbf{0.1516}\\ \hline
Colander & 0.2095 & \textbf{0.1249}\\ \hline
Cup & 0.2144 & \textbf{0.1430}\\ \hline
Pitcher & 0.2499 & \textbf{0.1131}\\ \hline
Saucepan & 0.1956 & \textbf{0.1103}\\ \hline
Scissors & 0.2391 & \textbf{0.1649}\\ \hline
Shaker & 0.2654 & \textbf{0.1453}\\ \hline
Thermos & 0.2271 & \textbf{0.1285}\\ \hline
\end{tabular}
}
\hfill
\parbox{0.5\linewidth} {
\caption{Mean 3D pose estimation error}
\label{table:2}
\centering
\begin{tabular}{*3c}
\hline
\textbf{Pose parameter} & \textbf{IRLS} & \textbf{OR-IRLS}\\ \hline
X (cm) & 1.6874 & \textbf{0.5774}\\ \hline
Y (cm) & 1.4953 & \textbf{0.6516}\\ \hline
Z (cm) & 8.228 & \textbf{2.1506}\\ \hline
Roll (degrees) & 1.1711 & \textbf{0.7152}\\ \hline
Pitch (degrees) & 7.9100 & \textbf{2.3191}\\ \hline
Yaw (degrees) & 5.7712 & \textbf{2.6055}\\ \hline
\end{tabular}
}
\end{table}
Figure~\ref{fig:3} shows that while both SD-HOP and~\citet{segaware} have similar recall at 1.0 FPPI, SD-HOP consistently preforms better in terms of area under the curve (AUC). Averaged over the 8 objects, SD-HOP achieves 16.13\% more AUC than~\citet{segaware}. Table~\ref{table:1} shows that SD-HOP consistently outperforms~\citet{segaware} in terms of segmentation error, achieving 42.44\% less segmentation error averaged over the 8 objects. Figure~\ref{fig:5} shows examples of the algorithm's output on various images from the CMU Kitchen Occlusion dataset.

\subsection{Ablation Study}
We conducted an ablation study on the `pitcher' object of the CMU Kitchen Occlusion dataset to determine the individual effect of our contributions. Using the loss function from~\citet{segaware} caused the segmentation error to increase from 0.1131 to 0.1547 and area under curve (AUC) of FPPI vs. recall to drop from 0.7877 to 0.7071. To discern the effect of 4-connected pairwise terms we removed the higher order terms from the model too. Using the pairwise terms as described in~\citet{segaware} caused the segmentation error to increase from 0.1547 to 0.2499 and AUC to decrease from 0.7071 to 0.6414.\par

Lastly, to quantify the effect of higher order potentials, we compared the full SD-HOP model against one with higher order potentials removed. Removing higher order potentials caused the segmentation error to increase from 0.1131 to 0.1430 and AUC to drop from 0.7877 to 0.7544. We hypothesize that for small objects like the ones in the CMU Kitchen Occlusion dataset, 4-connected pairwise terms are almost as informative as higher order terms. To check this hypothesis we tested the effect of removing higher order potentials on a close-up dataset of 41 images of a pasta-box occluded by various amounts through various household objects. Removing the higher order potentials caused the segmentation error to increase from 0.1308 to 0.1516 and area under curve AUC to drop from 0.9546 to 0.9008. This indicates that higher order terms are more useful for objects with larger and hence more informative segments.

\subsection{3D Pose Estimation} \label{subsec:3.3}
We collected 3D pose estimation results produced by IRLS and OR-IRLS on a dataset which has 17 images of a car-door in an indoor environment. The ground truth pose for the cardoor was obtained by an ALVAR marker~\citet{alvar_website}. Table~\ref{table:2} shows the mean errors in the six pose parameters. To discern the effect of errors inherent in the pose estimation process from the effect of occlusion reasoning, the pose of the cardoor was constant throughout the dataset, with various partial occlusions being introduced.\par

\begin{figure}[!t]
\centering
\subfigure{\includegraphics[width=0.24\textwidth]{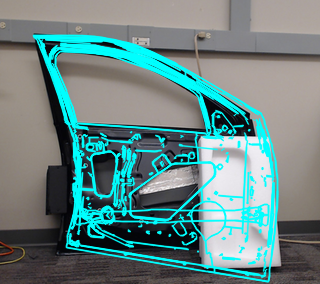}}
\subfigure{\includegraphics[width=0.24\textwidth]{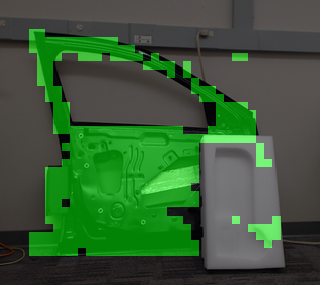}}
\subfigure{\includegraphics[width=0.24\textwidth]{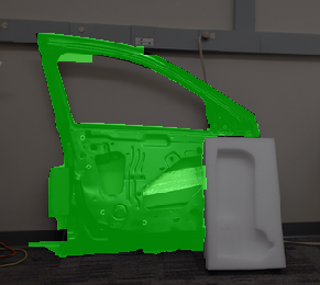}}
\subfigure{\includegraphics[width=0.24\textwidth]{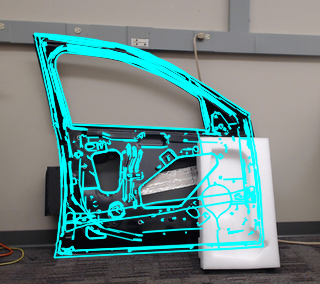}}
\caption{3D pose estimation. Left to right: Pose estimation with IRLS, SD-HOP raw segmentation mask, SD-HOP refined segmentation mask, Pose estimation with OR-IRLS. Best viewed in colour.}
\label{fig:4}
\end{figure}

The granular HOG cell-level mask produced by SD-HOP caused some important silhouette edges to be missed for pose estimation. To solve this problem we utilized the unsupervised segmentation done earlier for defining higher order terms. If more than 80\% of the area within a segment was marked 1, we marked the whole segment with 1. Since segments follow object boundaries, this produced much cleaner masks for pose estimation. Figure~\ref{fig:4} shows the masks and pose estimation results for an example image from the dataset, with more such examples presented in the supplementary material. Note that the segmentation errors mentioned in Table~\ref{table:1} use the raw masks.

\section{Conclusion} \label{sec:4}
We presented an algorithm (SD-HOP) that localizes partially occluded objects robustly and segments their visible regions accurately. In contrast to previous approaches that model occlusion, our algorithm uses higher order potentials to reason at the level of image segments and employs a loss function that targets both localization and segmentation performance. We demonstrated that our algorithm outperforms existing approaches on both tasks, when evaluated on a challenging dataset. Finally, we have shown that the segmentation output from SD-HOP can be used to improve pose estimation performance in the presence of occlusion. Avenues of future research include (1) training from weakly labelled data i.e. without segmentations, (2) a post-training algorithm to make object models comparable without having to train them together, and (3) using the occlusion information to reason about interactions between objects in scene understanding applications.\par

We would like to acknowledge Ana Huam{\'a}n Quispe's help with implementing this system on a bimanual robot. The system was used to enable the robot to pick up partially visible objects lying on a table.

\bibliography{occlusion}

\begin{thebibliography}{21}
\providecommand{\natexlab}[1]{#1}
\providecommand{\url}[1]{\texttt{#1}}
\expandafter\ifx\csname urlstyle\endcsname\relax
  \providecommand{\doi}[1]{doi: #1}\else
  \providecommand{\doi}{doi: \begingroup \urlstyle{rm}\Url}\fi

\bibitem[alv()]{alvar_website}
{ALVAR} tracking library.
\newblock
  \url{http://virtual.vtt.fi/virtual/proj2/multimedia/alvar/index.html}.
\newblock Acessed: 2015-05-03.

\bibitem[Arbelaez et~al.(2011)Arbelaez, Maire, Fowlkes, and Malik]{gpb_ucm}
Pablo Arbelaez, Michael Maire, Charless Fowlkes, and Jitendra Malik.
\newblock
  \href{http://ieeexplore.ieee.org/xpl/login.jsp?tp=&arnumber=5557884}{Contour
  detection and hierarchical image segmentation}.
\newblock \emph{Pattern Analysis and Machine Intelligence, IEEE Transactions
  on}, 33\penalty0 (5):\penalty0 898--916, 2011.
\newblock URL
  \url{http://ieeexplore.ieee.org/xpl/login.jsp?tp=&arnumber=5557884}.

\bibitem[Boykov and Kolmogorov(2004)]{kol_gc}
Yuri Boykov and Vladimir Kolmogorov.
\newblock
  \href{http://ieeexplore.ieee.org/xpls/abs_all.jsp?arnumber=1316848}{An
  experimental comparison of min-cut/max-flow algorithms for energy
  minimization in vision}.
\newblock \emph{Pattern Analysis and Machine Intelligence, IEEE Transactions
  on}, 26\penalty0 (9):\penalty0 1124--1137, 2004.
\newblock URL
  \url{http://ieeexplore.ieee.org/xpls/abs_all.jsp?arnumber=1316848}.

\bibitem[Boykov et~al.(2001)Boykov, Veksler, and Zabih]{boy_ae}
Yuri Boykov, Olga Veksler, and Ramin Zabih.
\newblock
  \href{http://ieeexplore.ieee.org/xpl/login.jsp?tp=&arnumber=969114}{Fast
  approximate energy minimization via graph cuts}.
\newblock \emph{Pattern Analysis and Machine Intelligence, IEEE Transactions
  on}, 23\penalty0 (11):\penalty0 1222--1239, 2001.
\newblock URL
  \url{http://ieeexplore.ieee.org/xpl/login.jsp?tp=&arnumber=969114}.

\bibitem[Choi and Christensen(2012)]{choi_pose}
Changhyun Choi and Henrik~I Christensen.
\newblock
  \href{http://ijr.sagepub.com/content/early/2012/03/01/0278364912437213}{Robust
  3D visual tracking using particle filtering on the special Euclidean group: A
  combined approach of keypoint and edge features}.
\newblock \emph{The International Journal of Robotics Research}, 31\penalty0
  (4):\penalty0 498--519, 2012.
\newblock URL
  \url{http://ijr.sagepub.com/content/early/2012/03/01/0278364912437213}.

\bibitem[Dalal and Triggs(2005)]{hog_main}
Navneet Dalal and Bill Triggs.
\newblock
  \href{http://ieeexplore.ieee.org/xpl/login.jsp?tp=&arnumber=1467360}{Histograms
  of oriented gradients for human detection}.
\newblock In \emph{Computer Vision and Pattern Recognition, 2005. CVPR 2005.
  IEEE Computer Society Conference on}, volume~1, pages 886--893. IEEE, 2005.
\newblock URL
  \url{http://ieeexplore.ieee.org/xpl/login.jsp?tp=&arnumber=1467360}.

\bibitem[Everingham et~al.(2010)Everingham, Van~Gool, Williams, Winn, and
  Zisserman]{pas_voc}
M.~Everingham, L.~Van~Gool, C.~K.~I. Williams, J.~Winn, and A.~Zisserman.
\newblock
  \href{http://link.springer.com/article/10.1007%2Fs11263-009-0275-4}{The
  Pascal Visual Object Classes (VOC) Challenge}.
\newblock \emph{International Journal of Computer Vision}, 88\penalty0
  (2):\penalty0 303--338, June 2010.
\newblock URL
  \url{http://link.springer.com/article/10.1007%2Fs11263-009-0275-4}.

\bibitem[Felzenszwalb and Huttenlocher(2004)]{fz_seg}
Pedro~F Felzenszwalb and Daniel~P Huttenlocher.
\newblock
  \href{http://link.springer.com/article/10.1023%2FB%3AVISI.0000022288.19776.77}{Efficient
  graph-based image segmentation}.
\newblock \emph{International Journal of Computer Vision}, 59\penalty0
  (2):\penalty0 167--181, 2004.
\newblock URL
  \url{http://link.springer.com/article/10.1023%2FB%3AVISI.0000022288.19776.77}.

\bibitem[Felzenszwalb et~al.(2010)Felzenszwalb, Girshick, McAllester, and
  Ramanan]{dpm_main}
Pedro~F Felzenszwalb, Ross~B Girshick, David McAllester, and Deva Ramanan.
\newblock
  \href{http://ieeexplore.ieee.org/xpls/abs_all.jsp?arnumber=5255236}{Object
  detection with discriminatively trained part-based models}.
\newblock \emph{Pattern Analysis and Machine Intelligence, IEEE Transactions
  on}, 32\penalty0 (9):\penalty0 1627--1645, 2010.
\newblock URL
  \url{http://ieeexplore.ieee.org/xpls/abs_all.jsp?arnumber=5255236}.

\bibitem[Gao et~al.(2011)Gao, Packer, and Koller]{segaware}
Tianshi Gao, Benjamin Packer, and Daphne Koller.
\newblock \href{http://ieeexplore.ieee.org/xpls/abs_all.jsp?arnumber=5995623}{A
  segmentation-aware object detection model with occlusion handling}.
\newblock In \emph{Computer Vision and Pattern Recognition (CVPR), 2011 IEEE
  Conference on}, pages 1361--1368. IEEE, 2011.
\newblock URL
  \url{http://ieeexplore.ieee.org/xpls/abs_all.jsp?arnumber=5995623}.

\bibitem[Girshick et~al.(2011)Girshick, Felzenszwalb, and
  Mcallester]{rbg_grammar}
Ross~B Girshick, Pedro~F Felzenszwalb, and David~A Mcallester.
\newblock
  \href{http://citeseerx.ist.psu.edu/viewdoc/summary?doi=10.1.1.231.2429}{Object
  detection with grammar models}.
\newblock In \emph{Advances in Neural Information Processing Systems}, pages
  442--450, 2011.
\newblock URL
  \url{http://citeseerx.ist.psu.edu/viewdoc/summary?doi=10.1.1.231.2429}.

\bibitem[Gould(2011)]{low_lin}
Stephen Gould.
\newblock
  \href{http://ieeexplore.ieee.org/xpl/articleDetails.jsp?arnumber=6945904}{Max-margin
  learning for lower linear envelope potentials in binary markov random
  fields}.
\newblock In \emph{Proceedings of the 28th International Conference on Machine
  Learning (ICML-11)}, pages 193--200, 2011.
\newblock URL
  \url{http://ieeexplore.ieee.org/xpl/articleDetails.jsp?arnumber=6945904}.

\bibitem[Hinterstoisser et~al.(2012)Hinterstoisser, Cagniart, Ilic, Sturm,
  Navab, Fua, and Lepetit]{line2d}
Stefan Hinterstoisser, Cedric Cagniart, Slobodan Ilic, Peter Sturm, Nassir
  Navab, Pascal Fua, and Vincent Lepetit.
\newblock
  \href{http://ieeexplore.ieee.org/xpls/abs_all.jsp?arnumber=6042881}{Gradient
  response maps for real-time detection of textureless objects}.
\newblock \emph{Pattern Analysis and Machine Intelligence, IEEE Transactions
  on}, 34\penalty0 (5):\penalty0 876--888, 2012.
\newblock URL
  \url{http://ieeexplore.ieee.org/xpls/abs_all.jsp?arnumber=6042881}.

\bibitem[Hsiao and Hebert(2012)]{cmu_occlusion}
Edward Hsiao and Martial Hebert.
\newblock
  \href{http://ieeexplore.ieee.org/xpl/login.jsp?tp=&arnumber=6248048}{Occlusion
  reasoning for object detection under arbitrary viewpoint}.
\newblock In \emph{Computer Vision and Pattern Recognition (CVPR), 2012 IEEE
  Conference on}, pages 3146--3153. IEEE, 2012.
\newblock URL
  \url{http://ieeexplore.ieee.org/xpl/login.jsp?tp=&arnumber=6248048}.

\bibitem[Joachims et~al.(2009)Joachims, Finley, and Yu]{joa_svm}
Thorsten Joachims, Thomas Finley, and Chun-Nam~John Yu.
\newblock
  \href{http://link.springer.com/article/10.1007%2Fs10994-009-5108-8}{Cutting-plane
  training of structural SVMs}.
\newblock \emph{Machine Learning}, 77\penalty0 (1):\penalty0 27--59, 2009.
\newblock URL
  \url{http://link.springer.com/article/10.1007%2Fs10994-009-5108-8}.

\bibitem[Kohli et~al.(2009)Kohli, Torr, et~al.]{kohli_hop}
Pushmeet Kohli, Philip~HS Torr, et~al.
\newblock
  \href{http://ieeexplore.ieee.org/xpl/login.jsp?tp=&arnumber=4587417}{Robust
  higher order potentials for enforcing label consistency}.
\newblock \emph{International Journal of Computer Vision}, 82\penalty0
  (3):\penalty0 302--324, 2009.
\newblock URL
  \url{http://ieeexplore.ieee.org/xpl/login.jsp?tp=&arnumber=4587417}.

\bibitem[Rother et~al.(2004)Rother, Kolmogorov, and Blake]{grab_cut}
Carsten Rother, Vladimir Kolmogorov, and Andrew Blake.
\newblock \href{http://dl.acm.org/citation.cfm?id=1015720}{Grabcut: Interactive
  foreground extraction using iterated graph cuts}.
\newblock In \emph{ACM Transactions on Graphics (TOG)}, volume~23, pages
  309--314. ACM, 2004.
\newblock URL \url{http://dl.acm.org/citation.cfm?id=1015720}.

\bibitem[Vedaldi and Zisserman(2009)]{vedaldi_structured}
Andrea Vedaldi and Andrew Zisserman.
\newblock Structured output regression for detection with partial truncation.
\newblock In \emph{Advances in neural information processing systems}, pages
  1928--1936, 2009.

\bibitem[Wang et~al.(2009)Wang, Han, and Yan]{wang_hog}
Xiaoyu Wang, Tony~X Han, and Shuicheng Yan.
\newblock
  \href{http://ieeexplore.ieee.org/xpl/login.jsp?tp=&arnumber=5459207}{An
  HOG-LBP human detector with partial occlusion handling}.
\newblock In \emph{Computer Vision, 2009 IEEE 12th International Conference
  on}, pages 32--39. IEEE, 2009.
\newblock URL
  \url{http://ieeexplore.ieee.org/xpl/login.jsp?tp=&arnumber=5459207}.

\bibitem[Xiang and Savarese(2013)]{sav_3Dasp}
Yu~Xiang and Silvio Savarese.
\newblock
  \href{http://ieeexplore.ieee.org/xpl/articleDetails.jsp?reload=true&arnumber=6755942}{Object
  Detection by 3D Aspectlets and Occlusion Reasoning}.
\newblock In \emph{Computer Vision Workshops (ICCVW), 2013 IEEE International
  Conference on}, pages 530--537. IEEE, 2013.
\newblock URL
  \url{http://ieeexplore.ieee.org/xpl/articleDetails.jsp?reload=true&arnumber=6755942}.

\bibitem[Zhu et~al.(2014)Zhu, Derpanis, Yang, Brahmbhatt, Zhang, Phillips,
  Lecce, and Daniilidis]{menglong}
Menglong Zhu, Konstantinos~G Derpanis, Yinfei Yang, Samarth Brahmbhatt, Mabel
  Zhang, Cody Phillips, Matthieu Lecce, and Kostas Daniilidis.
\newblock
  \href{http://www.cis.upenn.edu/~menglong/papers/icra2014_object_grasping.pdf}{Single
  image 3D object detection and pose estimation for grasping}.
\newblock In \emph{Robotics and Automation (ICRA), 2014 IEEE International
  Conference on}, pages 3936--3943. IEEE, 2014.
\newblock URL
  \url{http://www.cis.upenn.edu/~menglong/papers/icra2014_object_grasping.pdf}.

\end{thebibliography}
\end{document}